\documentclass[times, 10pt,twocolumn]{article} 
\usepackage{latex8}
\usepackage{times}
\usepackage{xcolor}
\usepackage{graphicx}
\usepackage{hyperref}
\usepackage{flushend}
\usepackage{amsmath,amssymb,amsfonts}

\begin{document}

\title{License Plate Privacy in Collaborative Visual Analysis of Traffic Scenes}

\author{Saeed Ranjbar Alvar, Korcan Uyanik, and Ivan V. Baji\'{c}\thanks{This work was supported in part by NSERC grants RGPIN-2021-02485 and RGPAS-2021-00038.}
\\
School of Engineering Science\\Simon Fraser University\\ Burnaby, BC, Canada\\ %Paolo.Ienne@di.epfl.ch\\
%\and
%First line of institution2 address\\ Second line of institution2 address\\ 
%SecondAuthor@institution2.com\\
}

\maketitle
\thispagestyle{empty}

\begin{abstract}
   Traffic scene analysis is important for emerging technologies such as smart traffic management and autonomous vehicles. However, such analysis also poses potential privacy threats. For example, a system that can recognize license plates may construct patterns of behavior of the corresponding vehicles' owners and use that for various illegal purposes. In this paper we present a system that enables traffic scene analysis while at the same time preserving license plate privacy. The system is based on a multi-task model whose latent space is selectively compressed depending on the amount of information the specific features carry about analysis tasks and private information.  Effectiveness of the proposed method is illustrated by experiments on the Cityscapes dataset, for which we also provide license plate annotations.    
\end{abstract}

%------------------------------------------------------------------------- 
\Section{Introduction}
\label{sec:introduction}

Traffic scene analysis is a key component of emerging technologies such as smart traffic management (STM) and autonomous vehicles (AV).
The goal of such analysis is to detect and recognize relevant objects such as vehicles, pedestrians, traffic signs, road/lane markers, etc., in order to make appropriate driving decisions in case of AV, or traffic management decisions in case of STM. Yet, such analysis also poses potential privacy threats. For example, a system that can recognize license plates may construct patterns of behavior of the corresponding vehicles' owners and use that for various malicious or illegal purposes. In fact, privacy in traffic monitoring and analysis has been studied for a while, but mostly in the context of localization and location-based services~\cite{Privacy_TMS_2006,Sherif_2017,Shahrom_2018,PbD_2021,traffic_density_2022}. Privacy in vision-based traffic analysis has been much less studied, and privacy related to license plates is a current topic of debate among privacy advocates, legal scholars, and lawmakers~\cite{ACLU_2013,ALPR_2020}. 

In this paper we present a system that enables visual traffic scene analysis while at the same time preserving license plate privacy. The system is based on a collaborative multi-task Deep Neural Network (DNN) whose latent space is selectively compressed depending on the amount of information the specific features carry about analysis tasks and private information. Applying a version of the information-theoretic privacy model called \emph{privacy fan}~\cite{VCIP_2021}, the amount of private information about license plates can be controlled via compression. 
Contributions of this paper include:
\vspace{-5pt}
\begin{itemize}
\itemsep0em 
    \item A multi-task collaborative model for analysis of traffic scenes incorporating license plate privacy. 
    \item Modified privacy fan that uses blurring to control privacy leakage. 
    \item License plate annotations for the Cityscapes validation set.
\end{itemize}

The paper is organized as follows.  Section~\ref{sec:preliminaries} presents preliminaries related to collaborative intelligence and the privacy fan.  Section~\ref{sec:proposed} presents the proposed methods, including the modified privacy fan. Experiments are presented in Section~\ref{sec:experiments} followed by conclusions in Section~\ref{sec:conclusions}.

\begin{figure*}[t]
\centering
    \includegraphics[width=0.8\textwidth]{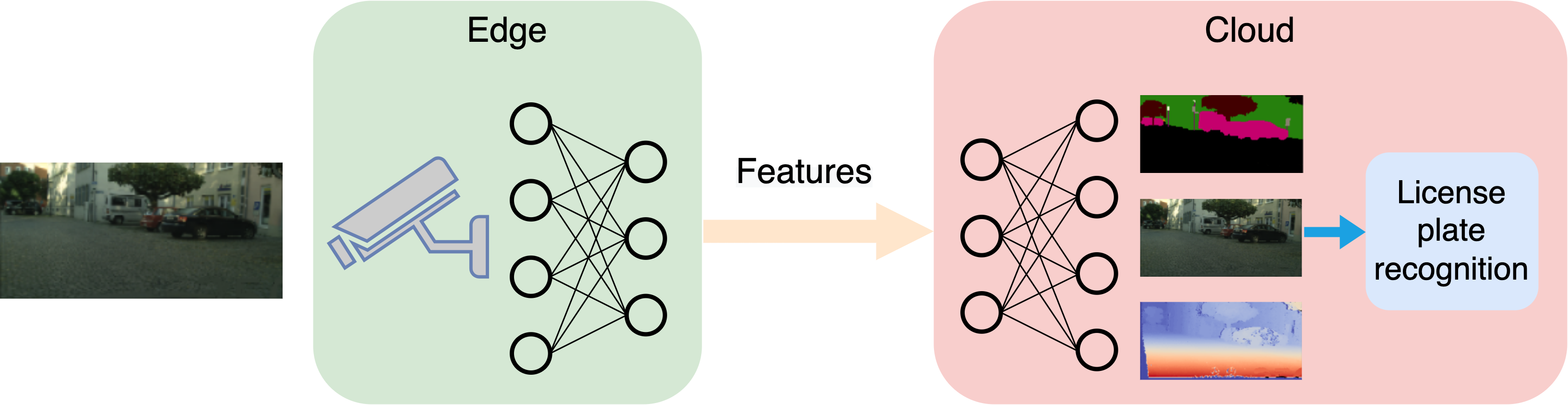}
   \caption{Collaborative multi-task  model with license plate privacy protection.}
   \label{fig:CI_model}
\end{figure*}

%------------------------------------------------------------------------- 
\Section{Preliminaries}
\label{sec:preliminaries}

\subsection{Collaborative intelligence}

Collaborative intelligence (CI)~\cite{CI_ICASSP_2021} is a way to deploy AI models across the edge and the cloud, to leverage resources of both. Typically, the front-end of the model runs on the edge device and sends the computed features to the cloud, where the back-end completes the inference. CI has been shown to have potential for energy savings at the edge, reduced inference latency~\cite{kang2017neurosurgeon,jointdnn}, as well as reduced bitrate (bandwidth) requirements~\cite{Hyomin_ICIP_2018,Hyomin_TIP_2022} for sending features from the edge to the cloud. Multi-task CI has also been a subject of recent research ~\cite{Saeed_ICIP_2019,Saeed_ICASSP_2020,Saeed_TIP_2021}, where the distributed CI model performs multiple tasks from the same set of features. The multi-task model from~\cite{Saeed_ICASSP_2020,Saeed_TIP_2021} forms the basis of the system presented in this paper.

\subsection{Privacy fan}

Privacy fan~\cite{VCIP_2021} is an information-theoretic privacy model ideally suited for CI. It assumes that the input $X$ is encoded into $C$ features $T_1$, $T_2$, ..., $T_C$, which will be used to support $L$ inference tasks $Y_1$, $Y_2$, ..., $Y_L$. In the context of CI, features are computed on the edge device and sent to the cloud to perform multiple tasks. Set  $\mathcal{P}$ contains indices of tasks that reveal private information. The goal is to select a subset $\mathcal{B}$ of at most $C'$ features that allow sufficient information for non-private tasks, while minimizing the information delivered to private tasks. The problem is formulated as~\cite{VCIP_2021}
\begin{equation}
\min_{\mathcal{B}:|\mathcal{B}|\leq C'} \;  \sum_{i\in\mathcal{B}}\sum_{j\in\mathcal{P}}I(T_i;Y_j), \quad
\textrm{s.t.} \;  \sum_{i\in\mathcal{B}}\sum_{j\notin\mathcal{P}}I(T_i;Y_j) \geq R,%\\
\label{eq:optimization_problem_2}
\end{equation}
where $I(T_i,Y_j)$ is the mutual information~\cite{Cover} between feature $T_i$ and task $Y_j$ and $R$ is the information rate of non-private tasks needed for sufficient accuracy. In the privacy fan~\cite{VCIP_2021}, features in $\mathcal{B}$ are only lightly compressed to preserve non-private information, while other features are more heavily compressed to remove private information.

\section{Proposed methods}
\label{sec:proposed}

The proposed collaborative multi-task model for traffic scene analysis is shown in Fig.~\ref{fig:CI_model}. Traffic scene image is captured at the edge device, where the front-end of the model computes the features and sends them to the cloud. The received features are then used for three tasks in the cloud: semantic segmentation, depth (disparity) estimation, and input image reconstruction. The 3-task model comes from~\cite{Saeed_ICASSP_2020,Saeed_TIP_2021}. The reconstructed image is then fed to the license plate recognition system from~\cite{ALPR_2018}, which first detects vehicles, then locates license plates, rectifies them, and then performs optical character recognition to read the plate. 

For license plate recognition to be successful, the input reconstruction back-end needs to be able to recover sufficient detail from the features received at the cloud. Our proposed method, to be described later in this section, selectively compresses the features produced by the front-end at the edge, such that details necessary for license plate recognition are removed, while the accuracy of semantic segmentation and depth estimation is preserved.

%\textcolor{blue}{ALPR system consists of 4 components~\cite{ALPR_2018}.  An image consisting of vehicles is inserted into YOLOv2 network which detects boundaries of vehicles. WPOD-NET is used to find the boundaries of license plates that have different obliqueness. The resulting image is rectified to transform into a rectangle and inserted to OCR-NET for character recognition.}

\subsection{Data}
We use the Cityscapes dataset~\cite{Cordts2016Cityscapes} to examine the performance of the proposed model. The 3-task model~\cite{Saeed_ICASSP_2020,Saeed_TIP_2021} that we build upon was trained on the 2975 images from the Cityscapes training set. We ran the license plate recognition model~\cite{ALPR_2018} on the 500 images from the Cityscapes validation set. Images where no license plates were detected were excluded from further analysis, while all images where a license plate was detected were manually inspected. There were 121 such images. For each license plate in those images, a bounding box was manually drawn and the license plate characters were manually recorded, if they were legible enough to the human eye. License plates that were too small or blurry to the human eye in the original image resolution of 2048$\times$1024 were denoted as unreadable. These manual annotations were used to examine recognition accuracy of the proposed system. We also make them publicly available.\footnote{\url{https://www.sfu.ca/~ibajic/datasets/Cityscapes\_val\_plates.zip}}

%\textcolor{red}{Korcan: please describe how you generated license plate annotations for the Cityscapes validation set. Also, please prepare these annotations for public release (one zip archive, with a README file). }
%\textcolor{blue}{The model has been run on validation dataset to find images that have no recognized license plates. From remaining, each image has been manually checked to see if a license plate image is visible. Bounding box locations of all license plates are manually extracted and normalized respect to the image resolutions. Visible license plates are annotated using the respective information and blurry license plates are annotated as dummies. (exclude from accuracy computation)}

\subsection{Blurring to prevent privacy leakage}
\label{sec:blurring}

\begin{figure}[t]
\centering
\includegraphics[width=0.4\textwidth]{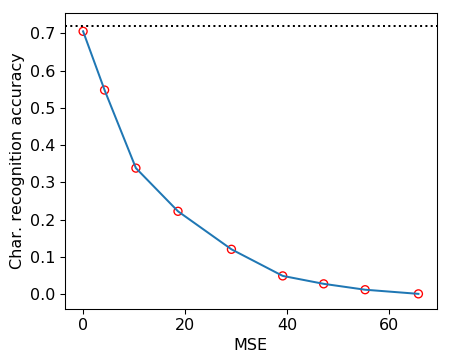}
\caption{Character recognition accuracy vs. MSE caused by Gaussian blurring. }
\label{fig:acc_vs_MSE}
\end{figure}

Recognizing the characters on the license plate requires sufficient detail and sharpness of the image, especially if the license plate is far from the camera and appears small. To confirm this observation, we blurred the images from the Cityscapes validation set using 11$\times$11 Gaussian filters with various levels of spread $\sigma$. The results of license plate character recognition on those images are shown in Fig.~\ref{fig:acc_vs_MSE} in terms of character recognition accuracy vs. the mean squared error (MSE) between the original and blurred image, averaged over the images in our set. As expected, recognition accuracy drops sharply with increasing blur.

This observation can be used to control privacy leakage related to license plates. By examining the latent space of the collaborative model from Fig.~\ref{fig:CI_model}, we noticed that removal of some features causes reconstructed images to lose fine detail, somewhat similar to Gaussian blurring. However, when removing or compressing features, we must ensure that the performance of the other analytics tasks is preserved. In the next section, we modify the privacy fan to incorporate this insight and arrive at a principled way of deciding how to compress the features in the latent space.  

\subsection{Modified privacy fan}

With Lagrangian relaxation, the privacy fan optimization problem can be reformulated~\cite{VCIP_2021} as
\begin{equation}
    \min_{\mathcal{B}:|\mathcal{B}|\leq C'} \sum_{i \in \mathcal{B}}\mathcal{L}_i,
\end{equation}
where the $i$-th Lagrangian term is given by
\begin{equation}
    \mathcal{L}_i = \sum_{j\in\mathcal{P}}I(T_i;Y_j) - \beta \cdot \sum_{j\notin\mathcal{P}}I(T_i;Y_j).
    \label{eq:i-th_lagrangian}
\end{equation}

For the model shown in Fig.~\ref{fig:CI_model}, let task 1 be semantic segmentation, task 2 be depth estimation, and task 3 be input reconstruction. We consider input reconstruction to be  privacy-revealing, hence $\mathcal{P}=\{3\}$, and the Lagrangian in~(\ref{eq:i-th_lagrangian}) becomes
\begin{equation}
\label{eq:largrangian_exp}
    \mathcal{L}_i = I(T_i;Y_3) - \beta \cdot \left[I(T_i;Y_1) +I(T_i;Y_2) \right].
\end{equation}
Based on the insight from Section~\ref{sec:blurring}, we replace the mutual information between a feature and the reconstructed input,  $I(T_i,Y_3)$, by the impact of that feature on the reconstructed input's MSE, $|\Delta\textsc{MSE}(T_i)|$. Specifically, we adopt the approach from~\cite{Molchanov_ICLR2017} to estimate the impact of the feature on the reconstruction MSE by computing the MSE between the output obtained by all features, $\widetilde{Y}_3$, and the output $\widetilde{Y}_3(T_i)$ obtained by zeroing out feature $T_i$. The new Lagrangian becomes
\begin{equation}
    \mathcal{L}_i = |\Delta\textsc{MSE}(T_i)| - \beta \cdot \left[I(T_i;Y_1) +I(T_i;Y_2) \right].
    \label{eq:new_lagrangian}
\end{equation}
Minimizing such a Lagrangian leads to selection of features that have minimal impact $|\Delta\textsc{MSE}(T_i)|$ on the input reconstruction, yet carry information about the other two tasks.

To create the base set $\mathcal{B}$ for the privacy fan, we sort the features according to their Lagrangian~(\ref{eq:new_lagrangian}), and then select the $C'$ features with minimum Lagrangian value. Features in $\mathcal{B}$ are only lightly compressed, while other features are more heavily compressed, as explained in the description of experiments below.
%According to~\cite{FRENAY2013}, MSE can be a proxy for mutual information, in the sense that maximizing mutual information minimizes MSE under certain conditions. In our case, we use this observation in reverse -- maximizing MSE to minimize mutual information -- hence the minus sign in front of $|\Delta\textsc{MSE}(T_i)|$ in~(\ref{eq:new_lagrangian}).

%------------------------------------------------------------------------- 

%------------------------------------------------------------------------- 
\section{Experiments}
\label{sec:experiments}
We used the 3-task DNN model from~\cite{Saeed_ICASSP_2020, Saeed_TIP_2021} in the experiments. The tasks performed by the 3-task model are: (1) semantic segmentation, (2) disparity map estimation, and (3) input reconstruction. The mentioned three tasks are selected due to the availability of sufficient amount of ground truth. The model was trained on the Cityscapes~\cite{Cordts2016Cityscapes} dataset using the 512 $\times$ 256 input images. Cross-entropy loss~\cite{metrics} and Mean Square Error (MSE) are used as the loss functions for semantic segmentation, and the other two tasks, respectively. The overall loss function is defined as: 
\begin{equation}
\label{eq:total_loss}
L =  \sum_{i=1}^{3} w_i L_i  + \log{\sqrt{\frac{1}{w_1}}} + \sum_{i=2}^{3} \log{\sqrt{\frac{1}{2 w_i}}} 
\end{equation}
where $L_i$ and $w_i$, $i=1,2,3$, are the task-specific losses and their weights, respectively. The weights $w_i$ are trainable parameters, and are updated during the training~\cite{cambridge}.

We asses task performance using mean Intersection over Union (mIoU)~\cite{metrics} for semantic segmentation and Root Mean Square Error (RMSE)~\cite{metrics} for disparity maps estimation. For license plate character recognition, the Character Recognition Accuracy (CRA) is defined as:
\begin{equation}
\label{eq:cra}
    \text{CRA} = (1 - \frac{\sum_{k=1}^{N} d(L_G^k, L_P^k)}{\text{total characters in the dataset}}) \times 100
\end{equation}
where $L_G^k$,$L_P^k$, $N$ and $d(.,.)$ are the ground-truth label for the $k$-th license plate, predicted label for the $k$-th license plate, the total number of plates in the dataset, and Levenshtein distance~\cite{levenshtein1966binary}, respectively.

The backbone of the multi-task model is similar to the backbone of YOLOv3~\cite{YOLOv3}. The first 37 layers of the backbone are processed on the edge. The remainder of the backbone in addition to the task-specific sub-models are processed on the cloud.   For an input $X \in \mathbb{R}^{H \times W \times 3}$ the edge outputs a feature tensor $T \in \mathbb{R}^{\frac{H}{8} \times \frac{W}{8} \times 256}$, i.e., the number of feature channels is $C= 256$. 

The features obtained from the edge ($T_i$) are sorted according to~\eqref{eq:new_lagrangian}. $|\Delta\textsc{MSE}(T_i)|$ is computed using the images in the training dataset. We adopted the method in~\cite{VCIP_2021} to compute the MI between the features and the desired output for semantic segmentation and the disparity map estimation.  The $C'$ features with minimum Lagrangian value are grouped as base features, and the rest are grouped as enhancement features. Following~\cite{VCIP_2021}, we select $C'= 179$. We also select $\beta=10$, to emphasize non-private tasks in~\eqref{eq:new_lagrangian}. 

Base and enhancement features are tiled into images, quantized using 8-bit min-max quantization and further compressed using HEVC RExt~\cite{HEVC_rext}. The base features are lossy encoded using QP=20, and the enhancement features are encoded using QP $\in\{40,30,20,10\}$. 

\begin{figure*}[tb!]
    \centering
    \begin{minipage}[b]{0.31\linewidth}
    \centering
    \includegraphics[width=\textwidth]{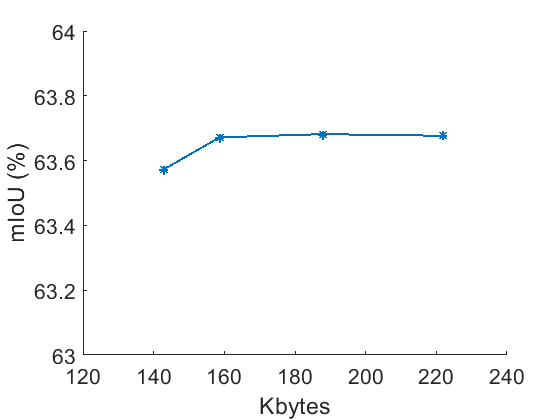}
    \centerline{(a)}\medskip
    \end{minipage}
    \begin{minipage}[b]{0.31\linewidth}
    \centering
    \includegraphics[width=\textwidth]{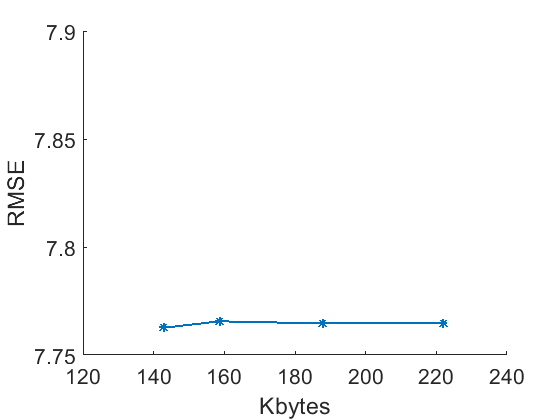}
    \centerline{(b)}\medskip
    \end{minipage}
    \begin{minipage}[b]{0.31\linewidth}
    \centering
    \includegraphics[width=\textwidth]{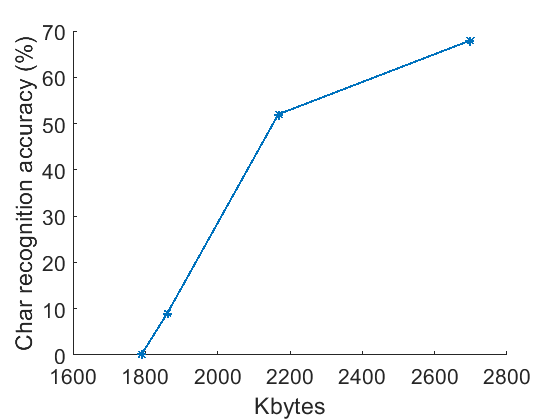}
    \centerline{(c)}\medskip
    \end{minipage}
    \hfill
\caption{Task accuracy vs. file size in Kbytes. (a) Semantic segmentation on $512\times256$ input images, (b) Disparity map estimation on $512\times256$ images (with RMSE, the lower, the better), (c) Character recognition accuracy on $2048\times1024$ images}
\label{fig:results_curves}
\end{figure*}

The task accuracies vs total file size for encoding the base and the enhancement features are shown in Fig.~\ref{fig:results_curves}. The four points correspond to encoding the base features with QP=20 and the enhancement features with QP $\in\{40,30,20,10\}$. It should be noted that the model was trained and tested on $512\times256$ images. However, with the resolution of  $512\times256$, license plates appear too small to be recognized using the system from~\cite{ALPR_2018}. Hence, to give the system a chance to perform recognition, the experiments for license plate recognition were performed on high-resolution images ($2048\times1024$). Hence, the total file size is larger on this task compared to the other two tasks.

As the graphs in the figure indicate, the performance of non-private tasks -- semantic segmentation and depth estimation -- remains roughly unchanged over the range of tested rates. However, the accuracy of license plate character recognition varies significantly over this range, from zero accuracy at the lowest rate (when the enhancement features are encoded with QP=40), to nearly 70\% accuracy at the highest rate (when the enhancement features are encoded using QP=10). This indicates that the private information needed to recognize license plates has indeed been placed in the enhancement features, and the amount of private information revealed can be controlled by adjusting the compression quality of the enhancement features. Meanwhile, base features have captured sufficient information to perform non-private tasks.

\begin{figure*}[tb!]
    \centering
    \begin{minipage}[b]{0.5\linewidth}
    \centering
    \includegraphics[width=\linewidth]{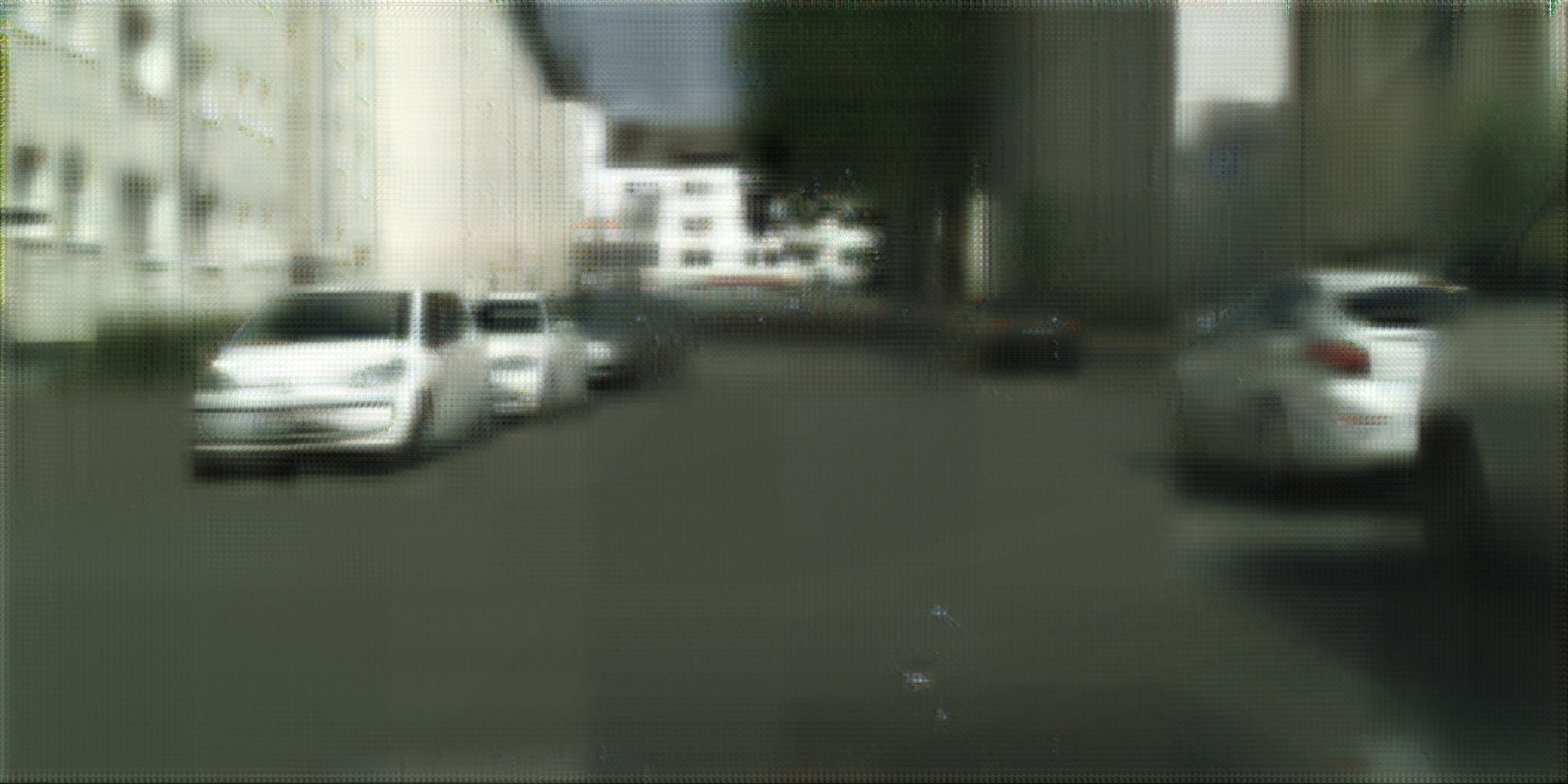}
    \centerline{(a)}\medskip
    \end{minipage}
    \begin{minipage}[b]{0.45\linewidth}
    \centering
    \includegraphics[width=0.5\linewidth]{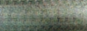}
    \vspace{60pt}
    \end{minipage}
    
    \centering
    \begin{minipage}[b]{0.5\linewidth}
    \centering
    \includegraphics[width=\linewidth]{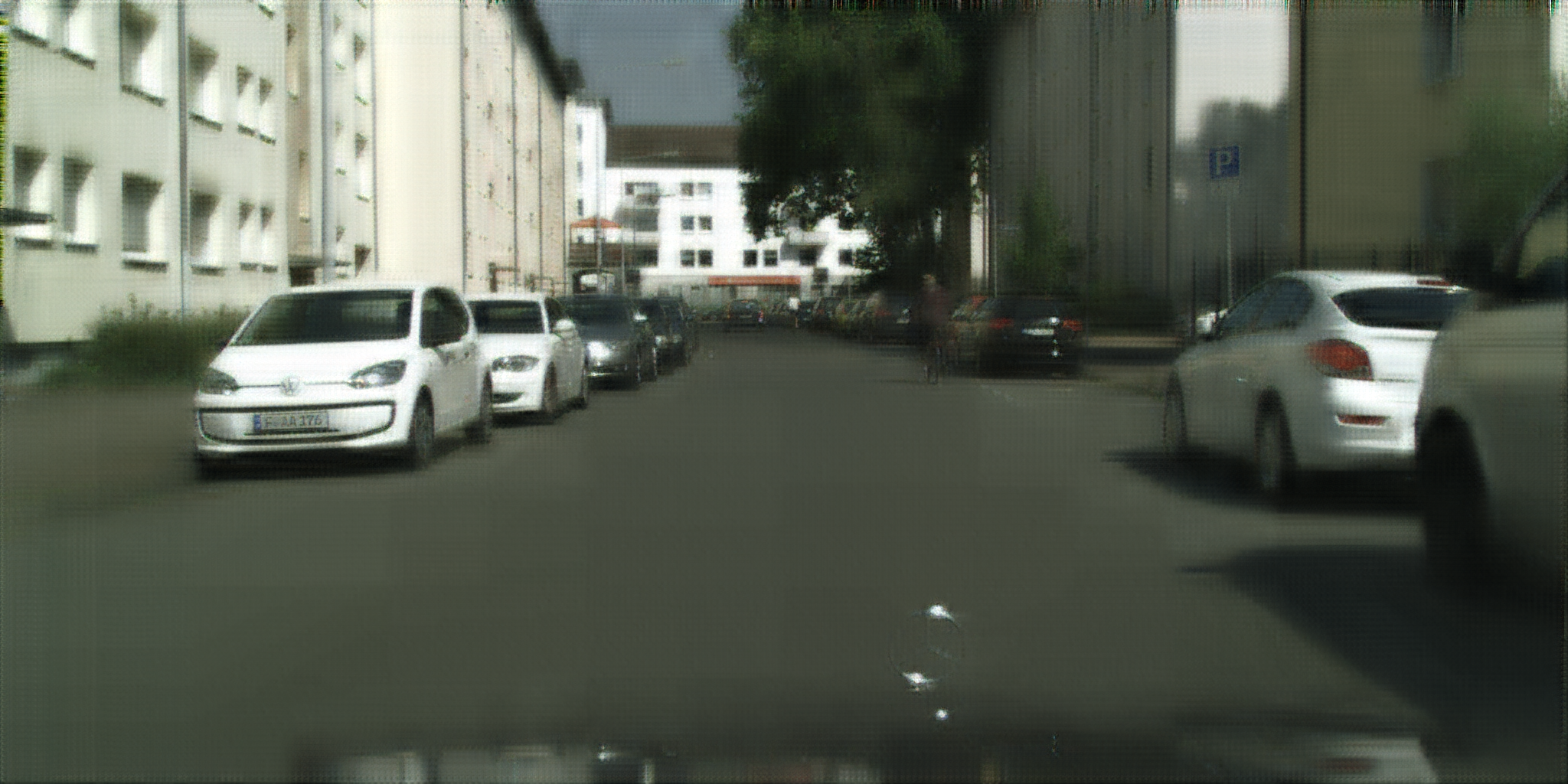}
    \centerline{(b)}\medskip
    \end{minipage}
    \begin{minipage}[b]{0.45\linewidth}
    \centering
    \includegraphics[width=0.5\linewidth]{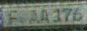}
     \vspace{60pt}
    %\centerline{(b)}\medskip
    \end{minipage}
    
     \centering
    \begin{minipage}[b]{0.5\linewidth}
    \centering
    \includegraphics[width=\linewidth]{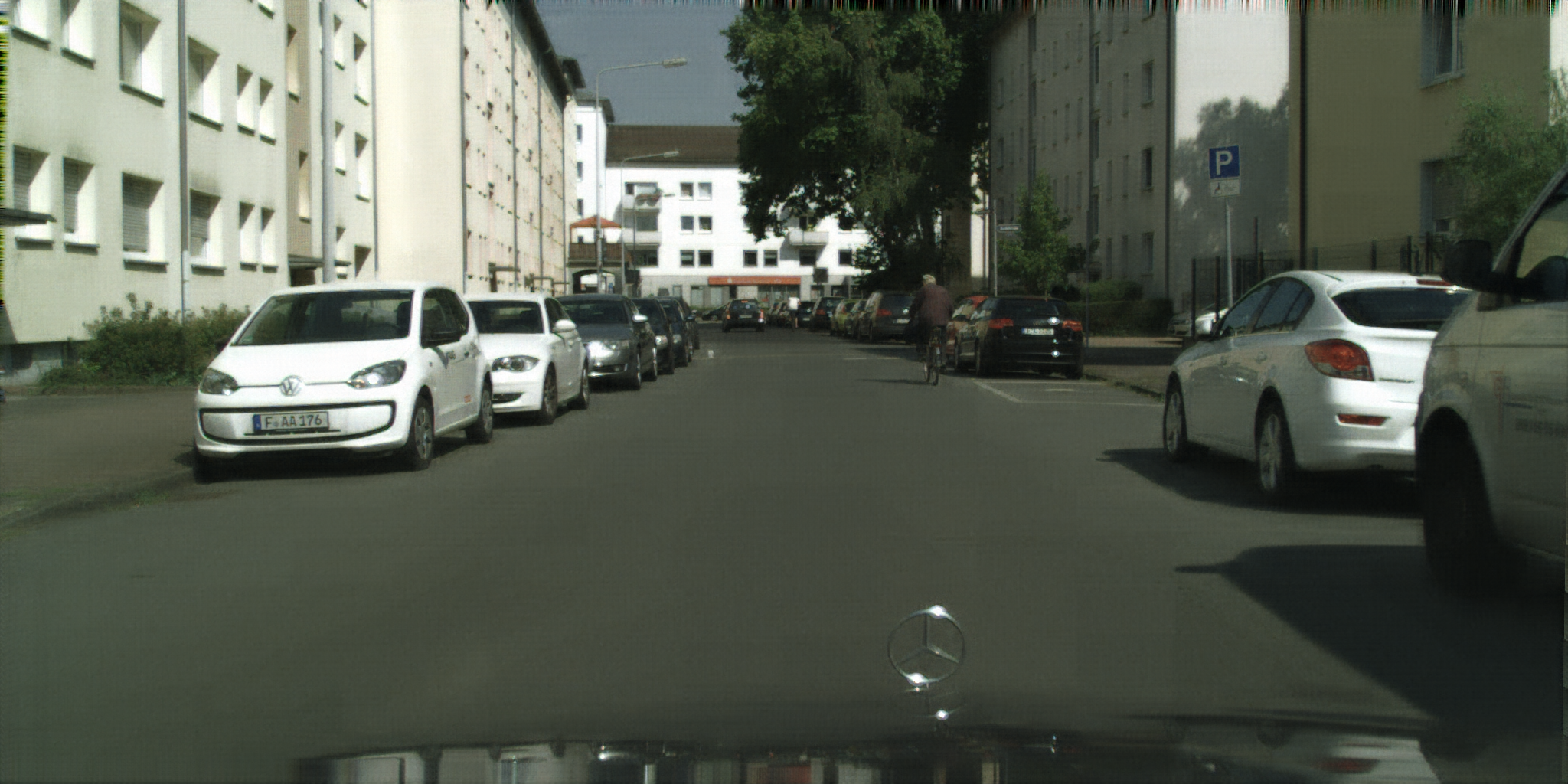}
    \centerline{(c)}\medskip
    \end{minipage}
    \begin{minipage}[b]{0.45\linewidth}
    \centering
    \includegraphics[width=0.5\linewidth]{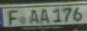}
     \vspace{60pt}
    %\centerline{(b)}\medskip
    \end{minipage}
    
    \centering
    \begin{minipage}[b]{0.5\linewidth}
    \centering
    \includegraphics[width=\linewidth]{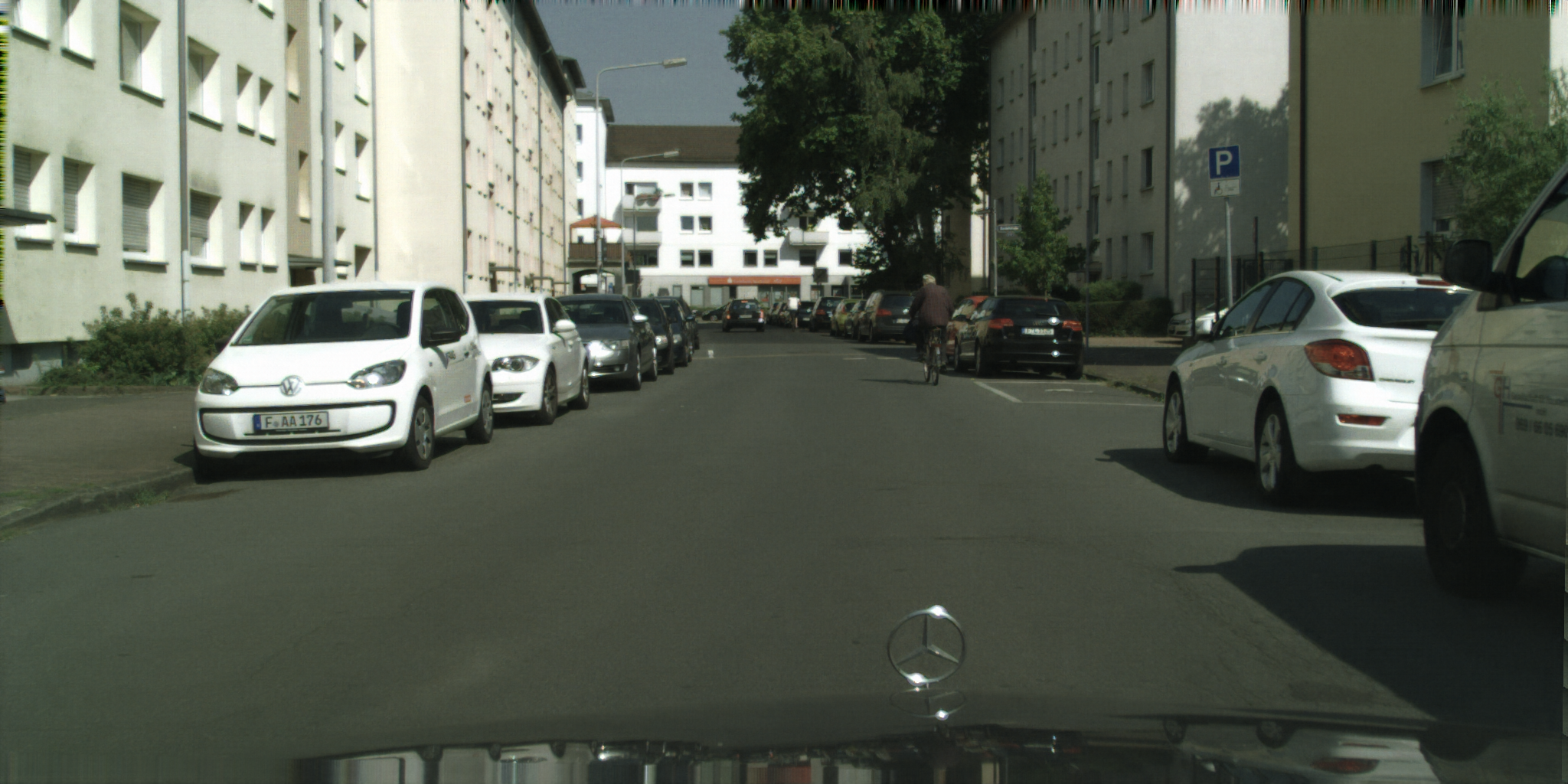}
    \centerline{(d)}\medskip
    \end{minipage}
    \begin{minipage}[b]{0.45\linewidth}
    \centering
    \includegraphics[width=0.5\linewidth]{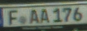}
     \vspace{60pt}
    %\centerline{(b)}\medskip
    \end{minipage}
    \hfill
\caption{A visual example of the reconstructed images for different QP values selected for encoding the enhancement features   (a) QP=40, (b) QP=30 , (c) QP=20, (d) QP=10. [Left] The reconstructed image, [Right]: The license plate in the reconstructed image}
\label{fig:rec_exp}
\end{figure*}

A visual example of the reconstructed input image obtained using the enhancement features encoded with different QP values is shown in Fig.~\ref{fig:rec_exp}. It is evident in the figure that the details of the reconstructed image are degraded as  QP corresponding to the enhancement features increases. As a result, the characters on license plate become less distinguishable as QP increases, and the license plate character recognition system fails to recognize the characters properly. Meanwhile, the performance of semantic segmentation and depth estimation remains roughly unchanged, as shown in Fig.~\ref{fig:results_curves}.

%------------------------------------------------------------------------- 
\section{Conclusions}
\label{sec:conclusions}

We presented a system for collaborative traffic scene analysis with license plate privacy protection. The system was built upon a 3-task model trained to perform semantic segmentation, depth estimation, and input reconstruction. Adapting the privacy fan approach to this scenario, and using the fact that loss of details can limit character recognition accuracy, we showed how to organize the latent space of the analysis model in such a way that the level of privacy (measured by character recognition accuracy) can be controlled via compression, while the accuracy of the other two tasks remains largely intact. The system was evaluated on the Cityscapes validation dataset, for which we also provided license plate annotations.

%------------------------------------------------------------------------- 
%\nocite{ex1,ex2}
\bibliographystyle{latex8}
\bibliography{refs}

\end{document}